\documentclass[letterpaper]{article} % DO NOT CHANGE THIS
\usepackage{aaai25}
\usepackage{adjustbox}% DO NOT CHANGE THIS
\usepackage{times}  % DO NOT CHANGE THIS
\usepackage{helvet}  % DO NOT CHANGE THIS
\usepackage{courier}  % DO NOT CHANGE THIS
\usepackage[hyphens]{url}  % DO NOT CHANGE THIS
\usepackage{graphicx} % DO NOT CHANGE THIS
\usepackage{natbib}  % DO NOT CHANGE THIS AND DO NOT ADD ANY OPTIONS TO IT
\usepackage{caption}

\usepackage{array}
\usepackage{algorithm}
\usepackage{algorithmic}

\usepackage{booktabs}
\usepackage{newfloat}
\usepackage{listings}

\DeclareCaptionStyle{ruled}{labelfont=normalfont,labelsep=colon,strut=off} % DO NOT CHANGE THIS
\floatstyle{ruled}
\newfloat{listing}{tb}{lst}{}
\floatname{listing}{Listing}
 \usepackage{color}
 \usepackage{amsmath}

\newcommand{\assistant}{{\texttt{AI Assistant in AEP}}}
\newcommand{\aepassistant}{{\texttt{AEP-Assistant}}}

\title{Evaluation and Incident Prevention in an Enterprise AI Assistant}
\author {
    Akash V. Maharaj\textsuperscript{\rm 1},
    David Arbour\textsuperscript{\rm 2}, 
    Daniel Lee\textsuperscript{\rm 1},
     Uttaran Bhattacharya\textsuperscript{\rm 2},
    \\
    Anup Rao\textsuperscript{\rm 2},
    Austin Zane\textsuperscript{\rm 3}, 
    Avi Feller\textsuperscript{\rm 3}, 
    Kun Qian\textsuperscript{\rm 1}, and 
    Yunyao Li\textsuperscript{\rm 1}
}
\affiliations {
    \textsuperscript{\rm 1}Adobe Inc.\\
    \textsuperscript{\rm 2}Adobe Research\\
    
    \textsuperscript{\rm 3}University of California, Berkeley\\
    maharaj@adobe.com, arbour@adobe.com, dlee1@adobe.com,  
    ubhattac@adobe.com, \\
    anuprao@adobe.com, austin.zane@berkeley.edu, afeller@berkeley.edu, kunq@adobe.com , yunyaol@adobe.com
}

\begin{document}

\maketitle

\begin{abstract}
Enterprise AI Assistants are increasingly deployed in domains where accuracy is paramount, making each erroneous output a potentially significant incident. This paper presents a comprehensive framework for monitoring, benchmarking, and continuously improving such complex, multi-component systems under active development by multiple teams. 
Our approach encompasses three key elements: (1) a hierarchical ``severity'' framework for incident detection that identifies and categorizes errors while attributing component-specific error rates, facilitating targeted improvements; (2) a scalable and principled methodology for benchmark construction, evaluation, and deployment, designed to accommodate multiple development teams, mitigate overfitting risks, and assess the downstream impact of system modifications; and (3) a continual improvement strategy leveraging multidimensional evaluation, enabling the identification and implementation of diverse enhancement opportunities. 
By adopting this holistic framework, organizations can systematically enhance the reliability and performance of their AI Assistants, ensuring their efficacy in critical enterprise environments. 
We conclude by discussing how this multifaceted evaluation approach opens avenues for various classes of enhancements, paving the way for more robust and trustworthy AI systems.
\end{abstract}

\section{Introduction}
The rapid growth and adoption of enterprise AI assistants have led to the emergence of complex, multi-agent systems that integrate various domain-specific capabilities \cite{generative_bot_ieee_2023,zaharia2024shift}. These \textit{compound} AI systems typically comprise several specialized \textit{agents}, each designed to address specific tasks such as question-answering over unstructured or structured data, and executing a particular set of actions \cite{kapoor2024ai}. While users interact with these multiple agents through a unified interface, developing each component often requires distinct technical approaches and, in large organizations, may be undertaken by separate teams.

This inherent complexity renders compound AI assistants susceptible to a spectrum of errors~\cite{hallucination_in_llms,openqa_eval,zhou2023don,gao2024llm}
These range from minor inaccuracies, easily identifiable and dismissible by users of varying expertise, to more insidious hallucinations that can challenge even experienced practitioners to detect. In certain domains, the latter category of errors may be classified as AI Incidents \cite{mcgregor2021preventing}, albeit with less public visibility than high-profile failures. Such errors can lead to immediate user frustration and, more critically, a long-term erosion of trust in the system \cite{choung2023trust, nourani2020role}.

The stakes are particularly high in enterprise scenarios, where AI assistants have the potential to significantly enhance worker productivity. However, errors in this context can result in wasted effort, flawed decision-making, or even more severe consequences \cite{levy2021assessing}. The delicate balance between leveraging AI's potential and mitigating its risks underscores the critical need for robust monitoring, evaluation, and improvement frameworks in enterprise AI deployments. 

Developing a trustworthy compound AI Assistant necessitates a high degree of sophistication in monitoring and evaluation, coupled with a systematic framework for continuous improvement. This paper presents an approach to addressing these challenges, drawing from experience in developing \assistant, a compound system designed to support marketers with varying levels of technical expertise and functions. \aepassistant~serves dual purposes: (1) answering conceptual questions using public production documentation as a knowledge base and (2) helping users interpret their proprietary operational data by querying a structured data store. This system has been deployed in production and was made generally available in June 2024 \cite{adobe_aep_assistant_ga_2024}.

Previous work \cite{maharaj-etal-2024-evaluation} introduced a preliminary approach to evaluation and continual improvement. This approach emphasized several key principles. First, it focused on improving \textit{leading} (rather than lagging) measures of user experience. Second, it implemented both component-wise and end-to-end evaluation. Third, it considered the entire user experience beyond the correctness of the ML models. Central to this approach was introducing a severity-based taxonomy for errors, coupled with a large-scale human annotation program applied to historical production traffic. This methodology was designed to meet our specific design goals while providing a comprehensive view of system performance. Notably, our previous work demonstrated that improvements in explainability and user experience (UX) enhancements could yield improvements comparable to those achieved through model upgrades. This finding underscores the multifaceted nature of AI system development and the importance of a holistic approach to improvement. It highlights that advancements in AI systems are not solely dependent on enhancing the underlying models, but also on making the system more interpretable and user-friendly \cite{ahn2021will, bayer2022role}.

Deployment and increasing scale can introduce new challenges that necessitate an evolution of this approach. Increasing production traffic requires the human annotation program to expand beyond a small group of expert annotators to a larger pool of non-experts. Additionally, a larger user base (beyond early testers) can bring decreased tolerance for regressions, compelling us to extend our evaluation framework to proactively estimate the impact of new features on error rates. Furthermore, scaling the \textit{development} of a multi-component system to multiple teams requires establishment of a repeatable set of evaluation best practices in order top prevent regressions and ensuring ongoing success.

In this paper, we recap and extend the prior work to address these new challenges. Our key contributions include:

\begin{itemize} 
    \item An in-depth analysis of the error severity taxonomy, presented as a comprehensive framework for the continuous detection and classification of AI incidents.
    \item A detailed guide to implementing and scaling this error severity-based framework, including description of a new sampling scheme that ensures a smaller set of data can be prioritized for annotation, and an ongoing adversarial testing program. 
    \item A thorough exploration of our approach to establishing representative shared evaluation datasets, including a stable long term holdout that enables proactive evaluation, and finally,
    \item An examination on how these holdouts can be used on an ongoing basis by multiple teams, and the expert-in-the-loop process used to develop and launch new features.
\end{itemize}

\newcommand{\ra}[1]{\renewcommand{\arraystretch}{#1}}

\begin{table}[t]
    \ra{1.1}  % Increase row height by 30%
    \centering
    \begin{tabular}{l | p{0.7\linewidth}}
        \toprule
        \textbf{Category} & \textbf{Definition} \\
        \midrule
        \textbf{Severity 0} & \textit{Answer looks right, but is wrong} \\
        \addlinespace
        \textbf{Severity 1} & \textit{Answer looks wrong, user can't recover} \\
        \addlinespace
        \textbf{Severity 2} & \textit{Answer looks wrong, user can recover} \\
        \bottomrule
    \end{tabular}
    \caption{Error Severity Framework for compound AI Assistants}
    \label{tab:errorseverity}
\end{table}

\section{Monitoring Quality in Enterprise AI Assistants}
The integration of AI into complex enterprise applications, such as marketing, offers significant potential benefits, including enhanced productivity and democratization of expertise. However, these advantages are accompanied by substantial challenges. Modern AI Assistants, powered by Large Language Models (LLMs) \cite{bommasani2021opportunities}, are known for their propensity towards fluency at the potential expense of accuracy and verifiability \cite{ji2023survey}. From a risk management perspective, each inaccurate response could be construed as a potential AI incident by a paying customer. 

It is important to recognize, however, that not all errors carry equal weight. The impact of an error can vary significantly depending on the specific use case. In some creative applications, a convincing LLM fabrication or hallucination might be tolerable or even desirable \cite{hubert2024current, jiang2024survey}. Conversely, in enterprise contexts where accuracy is paramount, such errors can be classified as incidents. Drawing from DevOps terminology \cite{kim2021devops}, we introduced the concept of:

\begin{itemize}
    \item \textbf{Severity-0} (``Sev-0'') errors: Categorizes instances where an answer is incorrect, yet the user lacks the ability to discern this inaccuracy. 
\end{itemize}

Less critical, but still significant, errors can also occur. We have extended to a tiered classification system to address these:

\begin{itemize}
    \item \textbf{Severity-1} (``Sev-1'') errors: These are characterized by clearly incorrect, irrelevant, or incoherent responses that users can identify, but where the user is unable to recover through query rephrasing. Examples include out-of-scope error messages or system errors. Such scenarios often lead to user frustration and potentially diminished trust in the system.
    \item \textbf{Severity-2} (``Sev-2'') errors: Similar to Sev-1 errors, these are identifiable by the user, but are typically recoverable through query rephrasing, allowing users to eventually obtain correct information.
\end{itemize}

This severity-based taxonomy as seen in Table \ref{tab:errorseverity} provides a structured framework for assessing and prioritizing errors in AI systems. In particular, where engineering and research teams are focused on reducing Severity-0 errors, with the goal of reducing these below 5\%.

\begin{figure}[t]
\centering
\includegraphics[width=\columnwidth]{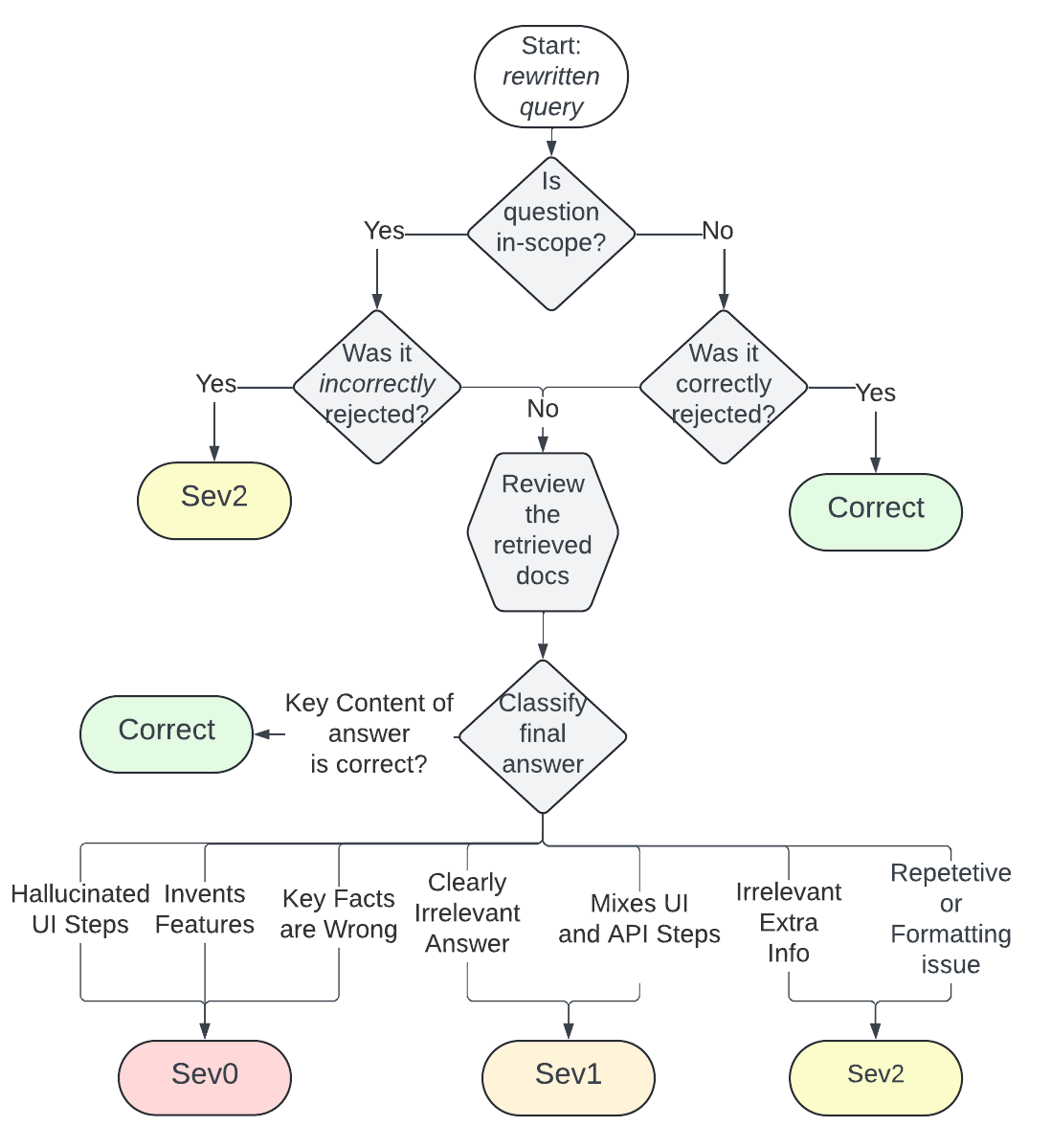} % 
\caption{A concrete implementation of how error severity is derived via a series of (more) objective human annotations}
\label{fig:errorseveritydecision}
\end{figure}

\subsection{Categorizing Errors by Severity}
While the error severity definitions presented in Table \ref{tab:errorseverity} are intuitively appealing, they inherently possess a degree of subjectivity that poses challenges in practical application. Several critical questions arise: How does one define the ``user" --- as an expert or a novice? How can we determine if a user actually recovers and obtains a correct answer during the course of the conversation? What criteria should be employed to assess whether something ``looks wrong"? These questions become increasingly pertinent when scaling an annotation program beyond a small group of expert annotators who are trained to have a unified understanding of the grading guidelines. These issues are analogous to existing problems faced in open-text and natural language generation human annotation/evaluation tasks \cite{van2021human}.

To address these challenges, we have developed a novel approach that decomposes the highly subjective criteria outlined in Table \ref{tab:errorseverity} into a series of less subjective human judgments. The error severity is then derived from these more granular assessments. Figure~\ref{fig:errorseveritydecision} illustrates the decision-making algorithm for determining error severity in the context of conceptual questions that require answers from public documentation. Each node in this decision tree represents either a human judgement to be made or a decision take by the compound AI system during the question-answering process. This decision tree translates into an annotation task comprising several less subjective questions for human annotators to answer. By breaking down the severity determination into these more objective decisions, we have significantly enhanced the robustness and stability of our human evaluation program. 

We acknowledge that some degree of subjectivity and disagreement is inevitable in such assessments. To mitigate this, we have implemented a system where expert annotators are called upon to resolve disagreements. This approach serves a dual purpose: it ensures the accuracy of the final annotation while also allowing for more efficient allocation of expert resources, which was a key objective of our original framework.

\subsection{Efficient Sampling for Human Annotation}
\begin{figure}[t]
\centering
\includegraphics[width=0.9\columnwidth]{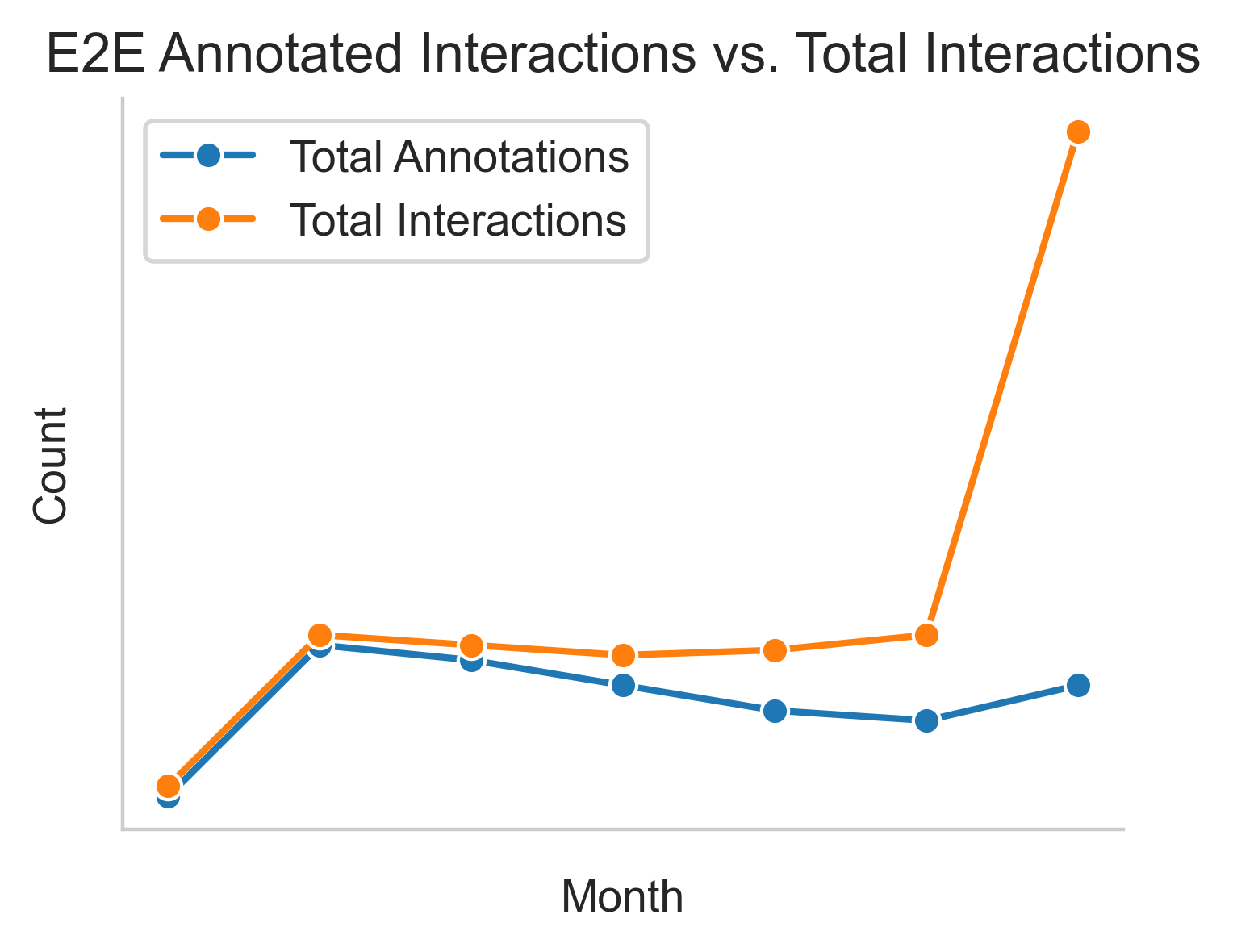}
\caption{An example of the number of annotations and interactions occurring after the public release of a compound AI Assistant. The specific $x$ and $y$ axis values have been omitted due to privacy concerns.}
\label{fig:interactgrowth}
\end{figure}

As compound AI Assistants become publicly available, there will naturally be a substantial increase in the volume of traffic, presenting new challenges in maintaining comprehensive quality assurance. This surge in interactions has rendered the approach of annotating every response unfeasible, as illustrated in Figure \ref{fig:interactgrowth}. The graph depicts the divergence between the total number of interactions and the number of annotated interactions over time, highlighting the growing gap between system usage and our ability to manually assess each output. While we have scaled our annotation efforts by expanding our team of annotators, this brute-force approach of augmenting labor has encountered both economic and logistical constraints. The challenge is further compounded when AI Assistants are used in a technical domain, which necessitates annotators to possess significant product expertise to provide accurate labels. This requirement for specialized knowledge creates a bottleneck in the annotation process, as the pool of qualified annotators is inherently limited. In response to these challenges, we have pivoted towards a strategy of subsampling interactions for annotation.

In addressing the subsampling challenge, the most straightforward and commonly applied solution is to draw query/answer pairs uniformly at random from the production dataset. While this approach is computationally simple, it can suffer from slow convergence rates to the true population error rate. 

However, given that we have access to additional information in the form of query/answer pairs, it is possible to achieve improved approximations by incorporating this ``covariate" information within the sampling process. This more sophisticated approach involves the construction of ``coresets", a concept that has gained traction in recent literature (\citet{feldman2020core,combettes2023revisiting,campbell2024general,ghadiri2024finite}).

Specifically, we assume that given queries, denoted $q$, and answers, denoted $a$ we assume that the quality (\textit{i.e.}, error severity) of an answer $a_i$ to query $q_i$ can be modeled as,  
$${quality} \sim f(\phi(q_i, a_i)) + \epsilon.$$ Here, $\phi$ is taken as a dense embedding representation of the query and answers, and $f$ is the space of linear functions~\footnote{Note that while this linearity assumption may appear restrictive, because of the choice of feature representation $\phi$, this is equivalent to training a final layer of a neural network after freezing previous layers.}.
Under this setting, choosing $k$ samples out of $N$ to minimize the approximation error is equivalent to solving a weighted discrepancy minimization problem, subject to an $L_0$ constraint~\citep{karnin2019discrepancy},
\begin{align}
\label{eq:disc}
    &\min_{\boldsymbol{w} \geq 0}\left\Vert\sum_i^N \boldsymbol{w}_i \phi(q_i, a_i) - \sum_i^N \phi(q_i, a_i)\right\Vert^2\\
    &\text{Subject to } \left\Vert\boldsymbol{w}\right\Vert_0 = k.
\end{align}
There are a host of algorithms that have been proposed as solutions to problem \ref{eq:disc} in various application contexts~(see, e.g., \citet{feldman2020core} and references therein). 
We were able to obtain the best results using greedy iterative geodesic ascent~(GIGA)~\citep{campbell2018bayesian}.

\begin{figure}[h!]
\centering
\includegraphics[width=0.8\columnwidth]{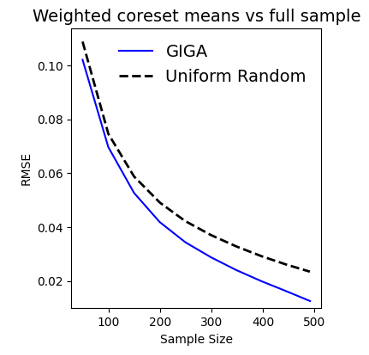} % Replace with your figure file path
\caption{Mean squared error of uniform (random) sampling and covariate aware sampling using GIGA. Error is measured with respect to proportion estimates obtained over the full set of annotations.}
\label{fig:rmse_sample}
\end{figure}

To measure the efficacy of coreset subsampling relative to uniform sampling we examined the ability of each to approximate the proportion of each error categories (as described in table \ref{tab:errorseverity}) in the full set of annotation. 
More concretely, defining $\hat{\theta}^k$ to be the proportions of $m$ error categories after sampling $k$ annotated interactions, and $\theta^n$ to be the proportions after observing the entire set of annotated interactions, we measure quality by considering the root mean squared error, $$\sqrt{\frac{1}{m}\sum_j^m (\hat{\theta}^k_j - \theta^n_j)^2},$$ for varying levels of $k$.
The experiments were conducted using 1000 boot strap iterations, where query/answer pairs and their accompanying annotations were drawn without replacement for each iteration.

For coreset based sampling we used llm2vec using LLama 3 8B instruct~\citep{behnamghader2024llm2vec} as an embedding model, after finding it to have the strongest empirical performance for the annotation sampling task, and the GIGA algorithm~(as described above) for sampling. 

Results can be seen in Figure \ref{fig:rmse_sample}, were we see a consistent advantage of the coreset based approach over random sampling, with the improvement increasing as a function of the number of samples being considered. 
Perhaps more importantly for annotation are the results shown in table 
\ref{tab:requiredsamples} which describes the number of samples that would have to be drawn using a uniform sampling strategy to reach the same RMSE as is obtained using coresets. 
The reduction in required samples across all sizes are non-trivial, ranging from 20 to 30 percent. Translated to hours of annotation, these improvements approximately correspond to a full day's worth of work for a trained expert annotator. 
\begin{table}[t]
\centering
\renewcommand{\arraystretch}{1.5}
\setlength{\tabcolsep}{8pt} % Adjust the cell padding
\begin{adjustbox}{max width=\columnwidth}
\begin{tabular}{|>{\centering\arraybackslash}p{2.5cm}|>{\centering\arraybackslash}p{2.5cm}|>{\centering\arraybackslash}p{2.5cm}|}
\hline
\textbf{Coreset size} & \textbf{Unif size} & \textbf{\% Reduction} \\
\hline
100 & 125 & 20\% \\
\hline
200 & 275 & 27\% \\
\hline
300 & 415 & 28\% \\
\hline
350 & 480 & 27\% \\
\hline
400 & 555 & 28\% \\
\hline
450 & 640 & 30\% \\
\hline
\end{tabular}
\end{adjustbox}
\caption{Coreset size vs Unif size with percentage reduction in the number of samples needed for uniform sampling to reach the equivalent root mean squared error of the coreset based approach.}
\label{tab:requiredsamples}
\end{table}

\subsection{Ongoing Adversarial Testing}
The implementation of an adversarial testing program represents a important strategy for monitoring and maintaining quality in AI-assisted systems. This approach involves engaging internal domain experts to systematically interact with the AI assistant with leaderboard gamification to improve engagement \cite{morschheuser2017gamified}, particularly focusing on complex queries. These experts are incentivized to identify and immediately flag erroneous responses, accompanied by comprehensive feedback. This feedback is instrumental in discerning the root causes of failures, which may stem from inadequacies in the retrieval process within the Retrieval-Augmented Generation (RAG) pipeline, instances of model hallucination, or gaps in the underlying documentation. The ongoing nature of this testing program proves invaluable for two primary reasons. Firstly, it allows for the prompt detection and categorization of system shortcomings, enabling targeted improvements. Secondly, the expert feedback serves as a rich source of annotated data, which can be leveraged for continuous model refinement and enhancement.

\section{Continual Improvement}
While the prior section focused on monitoring the quality of a compound AI system in production, this does not address the improvement process. Here, we discuss how shared evaluation datasets can be used for driving a more rigorous continual improvement process, and we also describe how the error severity framework is a multi-dimensional framework that motivates a holistic set of improvements. 

\begin{figure}[t]
\centering
\includegraphics[width=\columnwidth]{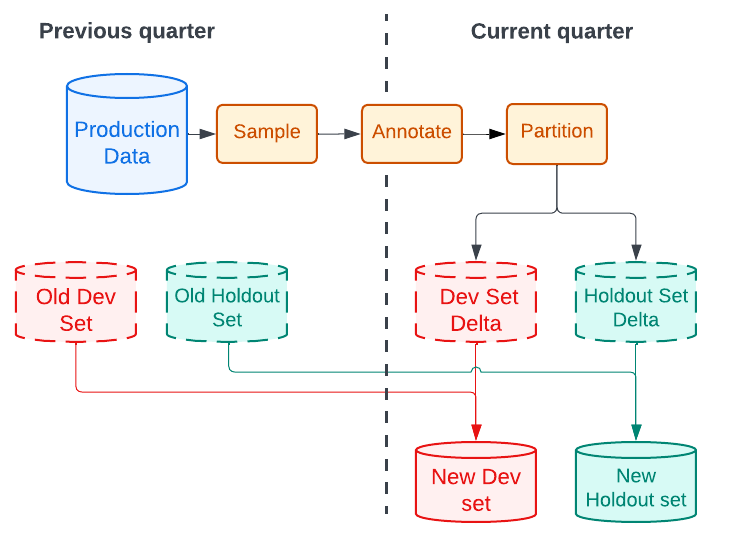} % Reduce the figure size so that it is slightly narrower than the column. Don't use precise values for figure width.This setup will avoid overfull boxes.
\caption{Creation of shared evaluation datasets on an ongoing basis, using sampling and human annotation of production traffic, which is then partitioned into development and holdout datasets. }
\label{fig:shared_eval}
\end{figure}

\subsection{Proactive Evaluation for Incident Prevention}
In the early phases of development, the primary focus is often on measuring and understanding \textit{historical} error rate, i.e., understanding how the system is behaving in production for early adopters. However, as more users on-board and a baseline user experience has been established, a new set of evaluation questions emerges whenever a new feature is proposed.

Specifically, for an incremental update to existing functionality, one is often interested in \textit{forecasting} how a feature will impact current end-to-end error rates. We emphasize end-to-end here: in contrast to an individual model developer who may care only about the performance of their model, as developers of the compound AI system, we are focused on end-to-end impact and the overall customer experience. Meanwhile, for brand new functionality, we are often interested in \textit{preventing regressions} of existing capabilities.  For both use cases, several secondary goals are also present: we would like to enable like-for-like comparisons over time as improvements are proposed, and would also like every team to use the same data for training and evaluation.

With these design goals in mind, we have developed a framework for shared evaluation datasets --- long-term development and holdout datasets based on sampled and gold-labeled production data. The development dataset can be used by any team developing new capabilities, as part of their regular development cycle. The holdout dataset is a long term test dataset that is not utilized by any team during component development, and is used purely for final evaluation before launching a new component. These datasets are refreshed on a quarterly basis, ensuring both stability over a sufficiently long period to enable the kinds of ``apples-to-apples'' comparisons we desire, but also allowing for responsiveness to changing customer behavior. 

The process of creating these datasets is illustrated in Figure~\ref{fig:shared_eval}. Near the end of each quarter, a representative subset of production data can be sampled using the same methods described earlier. This data, along with the production system decisions at the end of the quarter can be annotated by experts, and partitioned into a candidate ``development'' dataset, along with a smaller candidate ``holdout'' set. These candidate deltas are then merged with the prior period's development and holdout sets, to produce the new shared development and holdout sets. Note that the annotations here are comprehensive, representing all existing capabilities of the compound AI system. 

\begin{figure}[t]
\centering
\includegraphics[width=\columnwidth]{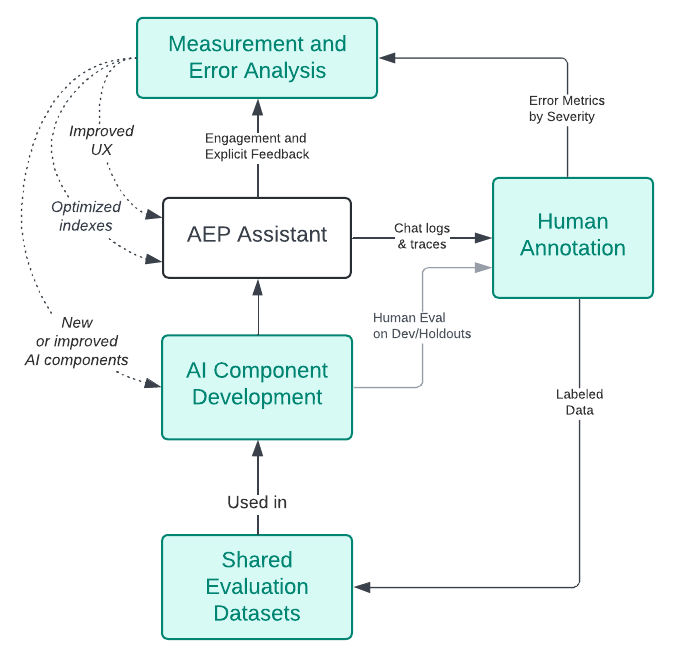} % Reduce the figure size so that it is slightly narrower than the column. Don't use precise values for figure width.This setup will avoid overfull boxes.
\caption{The continual improvement framework, emphasizing human annotation as a way of both generating labeled data to be used in shared evaluation datasets, and in driving measurement and error analysis. With the Error-severity framework, we are then able to prioritize improved AI components, but also consider other improvements like UX changes that aid in verifiability, explainability, and enhancing user's ability to recover.}
\label{fig:continuousimprovement}
\end{figure}

By using a sample derived from production data that is held out, one can compare the responses generated with proposed new functionality, against the existing production baseline. Having maintained the holdout, one can improve generalizability, and this along with the baseline comparison allows a forecast of the impact on our end-to-end error rates.  While several automated evaluation techniques can also be employed, including LLM-as-judge metrics \cite{zheng2024judging}, this evaluation may of course require a new round of human annotation, as illustrated in Figure~\ref{fig:continuousimprovement}. While expensive and slow, this cost should be balanced against the confidence gained with a true end-to-end assessment of impact, as opposed to an isolated component-wise accuracy improvement that is hard to judge. In a similar manner, for evaluating new functionality, one can ensure that no regressions are present by replaying queries and conversations from these shared development and holdout datasets, doing automated checks for regressions, and focusing human annotation effort on places where some change in an internal system decision, or response, has been identified. 

Finally, we note that the deployment of new features into production changes can change the baseline, and requires updating the benchmark labels. In this case, having done human evaluation against the holdout dataset proves to be valuable, as these new gold labels can be adopted as the new holdout dataset labels. The development set may however require larger human annotation exercises.

\subsection{Holistic System Improvement}

In alignment with \citet{kapoor2024ai}, we recognize that our Error Severity framework essentially constitutes a multidimensional evaluation paradigm. This approach prioritizes end-to-end error rates over the accuracies of individual components, compelling us to consider a broader spectrum of improvements rather than singularly focusing on optimizing against a specific domain benchmark.

This theory has been realized in practice, improving the users' mental model of the system, enhancing their ability to recover and ask different questions, providing interactive clarification, and improved explainability are all areas of active development. 

This framework is illustrated in Figure ~\ref{fig:continuousimprovement}, where we show an updated view of this continual improvement framework, emphasizing how human annotation produces the shared evaluation datasets used for capability development, but also contributes to error analysis. Our holistic approach to system evaluation and improvement integrates multiple data sources. We consider error rates derived from our severity framework, conversation metrics such as length, depth, and topic diversity, explicit user feedback in the form of ratings and comments, and implicit user feedback such as task completion rates and repeated queries. By synthesizing these diverse signals, we construct a comprehensive, data-driven roadmap for enhancing the system. This multifaceted strategy ensures that our improvements are not just focused on reducing errors, but on enhancing the overall user experience and system effectiveness.

\section{Conclusion and Future Work}
The development and maintenance of a flexible, scalable framework for monitoring and improving a domain-specific compound AI system presents a multitude of challenges. These challenges will evolve in tandem with the system's maturation, necessitating both scalable and repeatable annotation practices, as well as novel evaluation methodologies to forecast impact and prevent regressions. As scale expand, the number of contributing development teams can also grow, and we anticipate encountering new questions and challenges that will further test the robustness of this framework.

Our experience thus far has demonstrated the power of clearly defined metrics that are directly actionable by developers while remaining closely aligned with the user experience. These metrics serve as a potent organizing force, guiding our development efforts and quality assurance processes. Central to this approach has been our error severity framework, which has proven to be an invaluable tool in detecting critical errors and prioritizing remediation efforts.

As we continue to expand and refine our system, we envision broadening the scope of this framework to encompass a more comprehensive assessment of conversation quality. This expansion will move beyond mere end-to-end error rates to include nuanced aspects of user interaction such as:

\begin{itemize}
    \item \textbf{Consistency:} Coherence and relevance of responses throughout extended dialogues.
    \item \textbf{Conversational Experience:} Appropriateness of tone and style in different contexts.
    \item \textbf{Disambiguation:} Ability to handle ambiguity and seek clarification when needed.
    \item \textbf{Context:} Consistency of information provided across multiple interactions.
\end{itemize}

By evolving our framework in this direction, we aim to capture a more holistic view of the AI assistant's performance, ensuring that improvements in accuracy are balanced with enhancements in overall user experience and interaction quality.

While our current framework has proven effective in addressing the immediate challenges of monitoring and improving our domain-specific system, we acknowledge that the landscape of AI development and by extension, conversational support assistants is rapidly evolving. With it, needs an equally flexible methodology that can scale with our end-users and adapting system. 

\bibliography{IAAI-25-Evaluation-and-Benchmarking}

\end{document}